\documentclass[a4paper, english]{rnti} 
\usepackage[T1]{fontenc}
\usepackage[utf8]{inputenc}
\usepackage{url}
\usepackage{graphicx}
\usepackage{subcaption}
\usepackage{nccmath}
\usepackage{amsmath,amsfonts}
\usepackage{nccmath}
\usepackage[french]{babel}
\usepackage{listings}

\titrecourt{DEMAU: Decompose, Explore, Model \& Analyse Uncertainties}

\nomcourt{A. Hoarau, V. Lemaire}

\titre{DEMAU: Decompose, Explore, Model \& Analyse Uncertainties}

\auteur{Arthur Hoarau\affil{1}, Vincent Lemaire\affil{2}}

\affiliation{
    \affil{1}Université de Rennes, CNRS, IRISA, DRUID, France\\
    \affil{2}Orange innovation, Lannion, France
 }

\resume{
Recent research in machine learning has given rise to a flourishing literature on the quantification and decomposition of model uncertainty. This information can be very useful during interactions with the learner, such as in active learning or adaptive learning, and especially in uncertainty sampling. To allow a simple representation of these total, epistemic (reducible) and aleatoric (irreducible) uncertainties, we offer DEMAU, an open-source educational, exploratory and analytical tool allowing to visualize and explore several types of uncertainty for classification models in machine learning.
}

\summary{Recent research in machine learning has given rise to a flourishing literature on the quantification and decomposition of model uncertainty. This information can be very useful during interactions with the learner, such as in active learning or adaptive learning, and especially in uncertainty sampling. To allow a simple representation of these total, epistemic (reducible) and aleatoric (irreducible) uncertainties, we offer DEMAU, an open-source educational, exploratory and analytical tool allowing to visualize and explore several types of uncertainty for classification models in machine learning.}

\begin{document}


\section{Introduction}

In practical scenarios, the need to visualize uncertainty arises across various fields of machine learning, particularly during interactions with the learner, such as in active learning or adaptive learning, and especially in uncertainty sampling~\cite{Settles2009,Aggarwal2014,Hacohen2022}. It offers insights into the reliability of predictions and the confidence associated with model outputs. Commonly, uncertainty in probabilistic models has been represented using Shannon's entropy~\cite{Shannon1948}. However, the toolkit for uncertainty representation has expanded significantly. Researchers have proposed decomposing uncertainty into reducible (epistemic) and irreducible (aleatoric) components~\cite{Hora1996, eyke2019,kendall2017,senge2014,Charpentier2020}. Additionally, various frameworks for uncertainty reasoning have emerged~\cite{ABDAR2021243,huang2023review}, including probabilities, credal sets, possibilities, and belief functions~\citep{Dempster1967,shafer1976}.

The diversity of uncertainty forms and their implications necessitate effective visualization tools. Visualizing uncertainties can provide researchers and practitioners with deeper insights into model performance and decision-making processes. The interface proposed facilitates the exploration and interpretation of uncertainty in machine learning models, allowing users to comprehend the nuances of uncertainty representation across different frameworks.

Therefore, we propose DEMAU, an open-source educational, exploratory and analytical tool allowing to visualize and explore several types of uncertainty for classification models in machine learning, including recent techniques for computing epistemic and aleatoric uncertainties, while leaving the possibility for the user to customize the dataset, the model or even the method for quantifying uncertainty as they wish. The tool is mainly developed for research and educational purposes, and may not be suited for production use. It can be used to explore new datasets, to analyze or develop uncertainty quantification metrics or even to visualize and decompose the uncertainty of a machine learning model. It may also be used to facilitate the undertanding for newcomers in the field of uncertainty quantification, such as PhD students.

\begin{figure}
    \centering
    \includegraphics[width=0.99\linewidth]{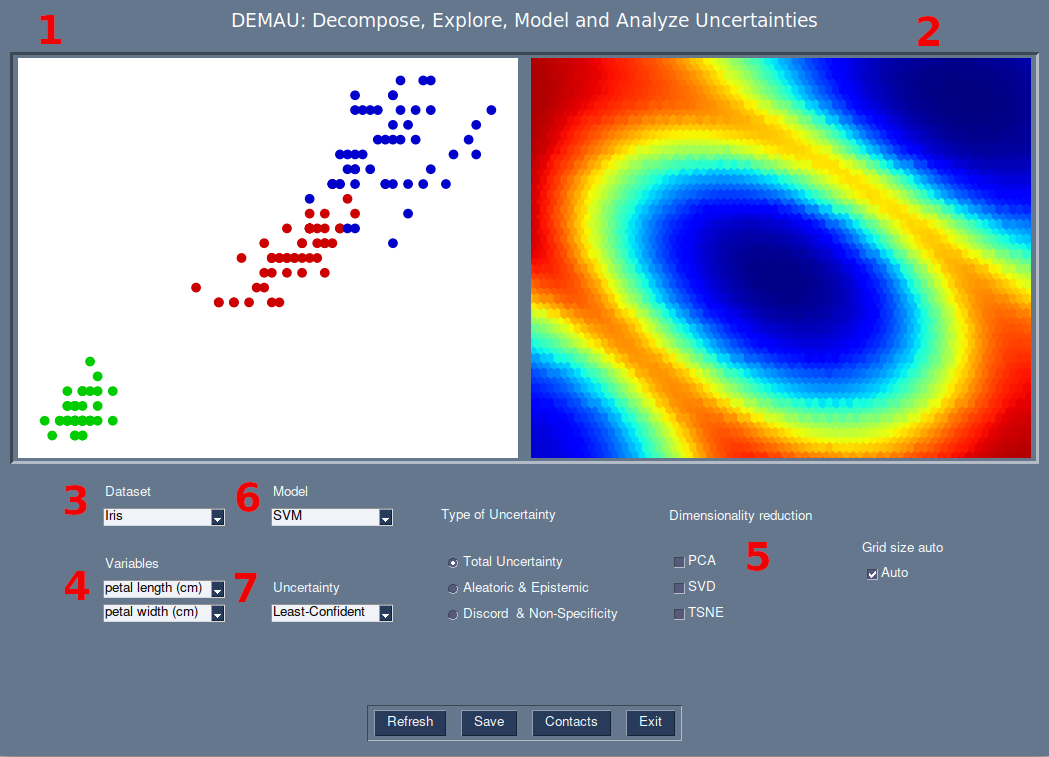}
    \caption{DEMAU interface}\label{fig:DEMAU}
\end{figure}

\section{DEMAU to explore and decompose uncertainties}

\subsection{Overview}

The interface is presented in Figure~\ref{fig:DEMAU}, with two main panels including the dataset currently studied (in panel (1)) along with the areas of uncertainty (in panel (2)) with a gradient from red to blue. More red indicates an area of high uncertainty for the model while a blue area indicates high certainty in the prediction. With the drop-down menu (3) the user can select one of the pre-loaded datasets, or use its own (see section~\ref{sec:mod}). In the double drop-down menu (4) two variables, among the ones in the dataset, can be selected for a 2D representation, along with reduction techniques, according to the checkboxes (5). DEMAU also comes with different possible machine learning models in the drop-down menu (6) (see section~\ref{sec:models}) and several uncertainties in the drop-down menu (7) can be computed (see section~\ref{sec:methods} and~\ref{sec:decompo}). In addition, you can simply add your own datasets, models and uncertainty quantification methods (see section~\ref{sec:mod}).

\subsection{Uncertainty from machine learning models}\label{sec:models}

\begin{figure}
    \centering
    \begin{subfigure}{.24\linewidth}
    \includegraphics[width=\linewidth]{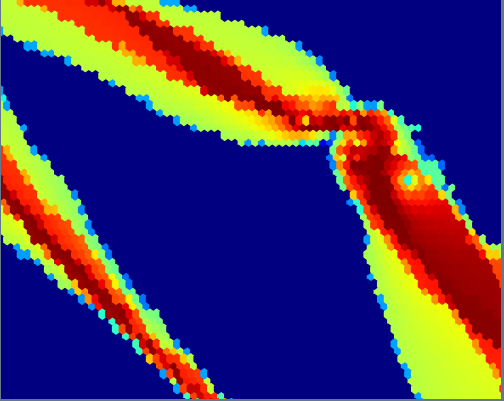}
    \caption{\scriptsize K-NN}\label{fig:knn}
    \end{subfigure}
    \begin{subfigure}{.24\linewidth}
    \includegraphics[width=\linewidth]{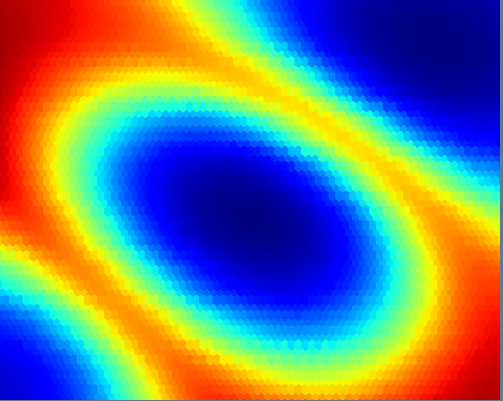}
    \caption{\scriptsize SVM}\label{fig:svm}
    \end{subfigure}
    \begin{subfigure}{.24\linewidth}
    \includegraphics[width=\linewidth]{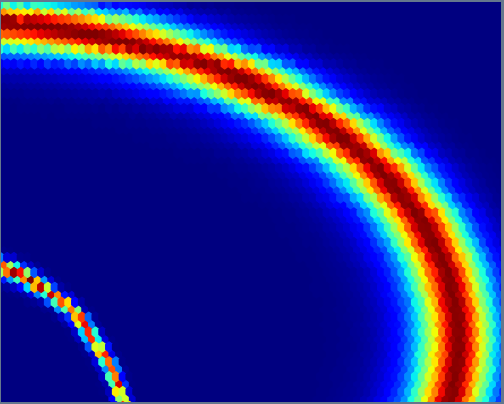}
    \caption{\scriptsize Naive Bayes}\label{fig:nb}
    \end{subfigure}
    \begin{subfigure}{.24\linewidth}
    \includegraphics[width=\linewidth]{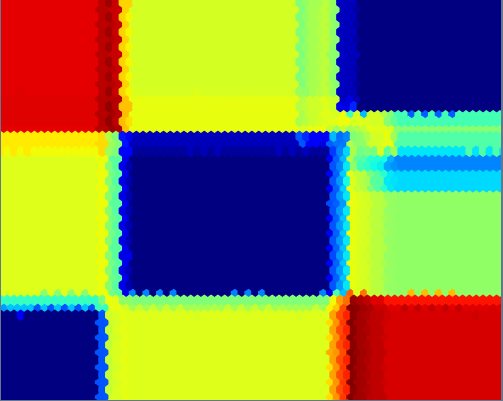}
    \caption{\scriptsize Random Forest}\label{fig:rf}
    \end{subfigure}
    \caption{Uncertainty for several models.}\label{fig:models}
\end{figure}

DEMAU simply allows to represent areas of uncertainty from your model. Figure~\ref{fig:models} shows different areas of uncertainty for several machine learning models according to Fisher's Iris dataset, presented in Figure~\ref{fig:DEMAU}. Built-in models are the K-Nearest Neighbors~(\ref{fig:knn}), Support Vector Machine~(\ref{fig:svm}), Naive Bayes classifier~(\ref{fig:nb}) and Random Forest~(\ref{fig:rf}), but the user is free to use its own model.

\subsection{Use of different methods}\label{sec:methods}

\begin{figure}
    \centering
    \begin{subfigure}{.24\linewidth}
    \includegraphics[width=\linewidth]{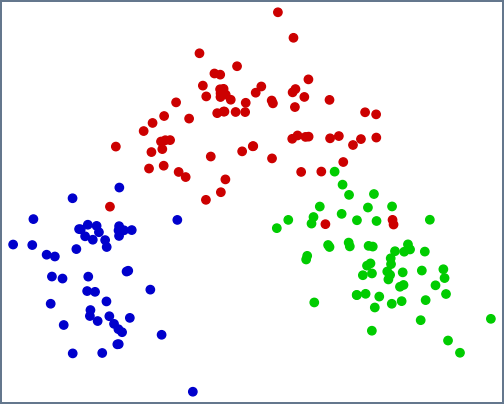}
    \caption{\scriptsize Dataset}
    \end{subfigure}
    \begin{subfigure}{.24\linewidth}
    \includegraphics[width=\linewidth]{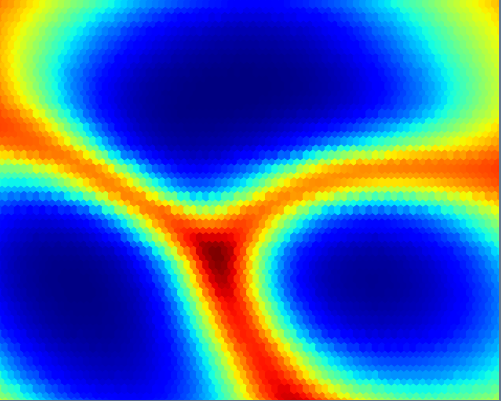}
    \caption{\scriptsize Gini}\label{fig:gini}
    \end{subfigure}
    \begin{subfigure}{.24\linewidth}
    \includegraphics[width=\linewidth]{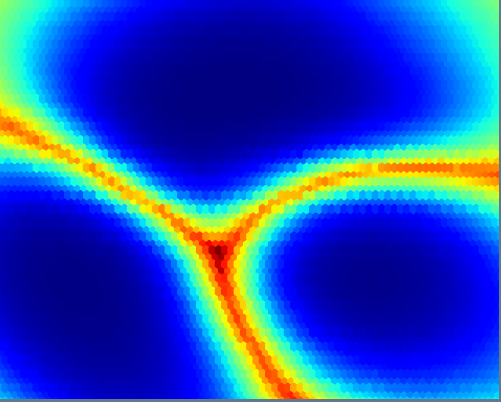}
    \caption{\scriptsize Entropy}\label{fig:entropy}
    \end{subfigure}
    \begin{subfigure}{.24\linewidth}
    \includegraphics[width=\linewidth]{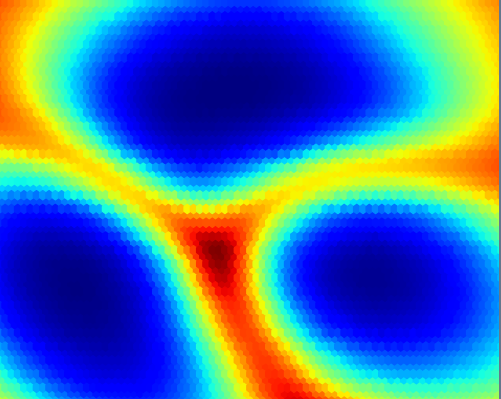}
    \caption{\scriptsize Least Confident}\label{fig:least-confident}
    \end{subfigure}
    \caption{Different uncertainty quantifications.}\label{fig:quant}
\end{figure}

Figure~\ref{fig:quant} shows different computations of the model uncertainty, with Gini criteria~(\ref{fig:gini}), Shannon's entropy~(\ref{fig:entropy}) and a Least-Confident measure~(\ref{fig:least-confident}). Although these measurements appear to represent similar areas, they may vary and any other measurement can also be simply added for different results.

\subsection{Decomposition of uncertainty}\label{sec:decompo}

\begin{figure}
    \centering
    \begin{subfigure}{.24\linewidth}
    \includegraphics[width=\linewidth]{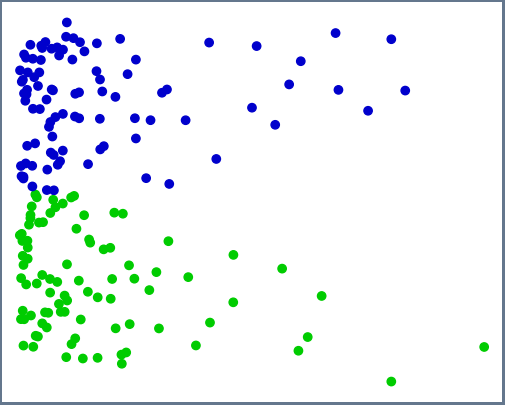}
    \caption{\scriptsize Dataset}
    \end{subfigure}
    \begin{subfigure}{.24\linewidth}
    \includegraphics[width=\linewidth]{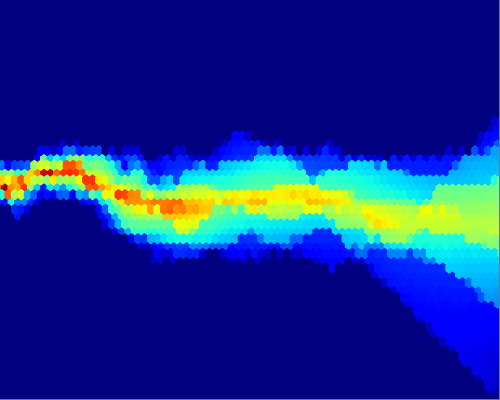}
    \caption{\scriptsize Aleatoric}\label{fig:aleatoric}
    \end{subfigure}
    \begin{subfigure}{.24\linewidth}
    \includegraphics[width=\linewidth]{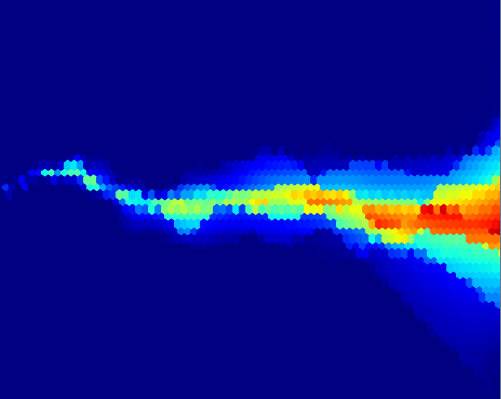}
    \caption{\scriptsize Epistemic}\label{fig:epistemic}
    \end{subfigure}
    \begin{subfigure}{.24\linewidth}
    \includegraphics[width=\linewidth]{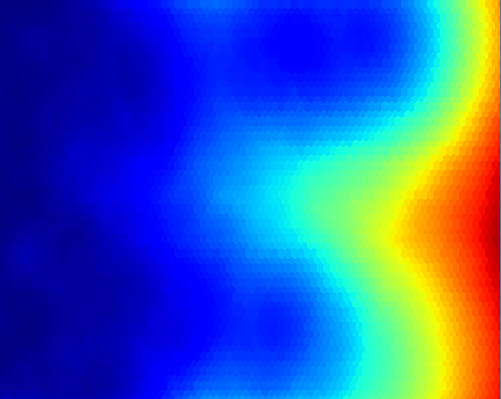}
    \caption{\scriptsize Non-Specicifity}\label{fig:nonspe}
    \end{subfigure}
    \caption{Decomposition of uncertaintiy.}\label{fig:decomp}
\end{figure}

Recent methods for estimating the uncertainty of the model rely on the decomposition into reducible and irreducible uncertainties. Figure~\ref{fig:decomp} shows the areas of uncertaintiy corresponding to three different methods. Built-in computations allow to represent an estimation of the aleatroic uncertainty~(\ref{fig:aleatoric}), the epistemic uncertainty~(\ref{fig:epistemic}), the non-specificity~(\ref{fig:nonspe}) and the discord of the model. Note that the bibliographic references associated with each method are directly displayed within the interface.

\subsection{Modularity}\label{sec:mod}

New dataset can be added by simply deposing a CSV file into the \emph{datasets} folder. It is not needed to reload the app, just press the \emph{refresh} button. Machine learning models can be added in the \emph{models} file and uncertainty quantification methods in the \emph{uncertainties} file.
A built-in fuction also allows to reduce the grid size for faster computation on very large datasets, light configurations or greedy strategies.

\section{Conclusion}

By offering a user-friendly interface, DEMAU empowers researchers and practitioners to gain deeper insights into the uncertainties associated with their models, thereby enhancing decision-making processes and model performance assessment. This information can be very useful during interactions with the learner, such as in active learning or adaptive learning, and especially in uncertainty sampling. The modular design of DEMAU allows for easy customization and extension, facilitating its utility across various applications and datasets. Moving forward, DEMAU serves as a valuable tool for advancing research and education in uncertainty quantification, while also promoting transparency and reproducibility in machine learning experiments.\\

\noindent{\bf Access:} DEMAU is an open-source Python software for educational, exploratory and analytical purposes allowing to visualize and explore several types of uncertainty for classification models in machine learning. Link to the project: \url{https://github.com/ArthurHoa/DEMAU}. \\A presentation video of DEMAU is available at: \url{https://youtu.be/2xHwqlJJTm4}

\Fr
\bibliographystyle{rnti}
\bibliography{ref/egc}

\end{document}